\documentclass[letterpaper]{article} % DO NOT CHANGE THIS
\usepackage[]{aaai23}  % DO NOT CHANGE THIS
\usepackage{times}  % DO NOT CHANGE THIS
\usepackage{helvet}  % DO NOT CHANGE THIS
\usepackage{courier}  % DO NOT CHANGE THIS
\usepackage[hyphens]{url}  % DO NOT CHANGE THIS
\usepackage{graphicx} % DO NOT CHANGE THIS
\usepackage{natbib}  % DO NOT CHANGE THIS AND DO NOT ADD ANY OPTIONS TO IT
\usepackage{caption} % DO NOT CHANGE THIS AND DO NOT ADD ANY OPTIONS TO IT
\usepackage{algorithm}
\usepackage{algorithmic}

\usepackage{dsfont}
\usepackage{booktabs}
\usepackage{multirow}
\usepackage{makecell}
\usepackage{xspace}
\usepackage{xcolor}
\usepackage{amsmath}
\usepackage{amssymb}
\usepackage{enumitem}
\usepackage{subfigure}
\usepackage{bm}
\usepackage{newfloat}
\usepackage{listings}
\DeclareCaptionStyle{ruled}{labelfont=normalfont,labelsep=colon,strut=off} % DO NOT CHANGE THIS
\floatstyle{ruled}
\newfloat{listing}{tb}{lst}{}
\floatname{listing}{Listing}
\newcommand{\myparagraph}[1]{\vspace{3.0pt}\noindent{\bf #1}}

\makeatletter
\DeclareRobustCommand\onedot{\futurelet\@let@token\@onedot}
\def\@onedot{\ifx\@let@token.\else.\null\fi\xspace}

\renewcommand*{\@fnsymbol}[1]{\ensuremath{\ifcase#1\or \dagger\or *\or \ddagger\or
   \mathsection\or \mathparagraph\or \|\or **\or \dagger\dagger
   \or \ddagger\ddagger \else\@ctrerr\fi}}

\makeatother

\title{Fair Visual Recognition via Intervention with Proxy Features}
\author{
    Yi Zhang,\textsuperscript{\rm 1}
    Jitao Sang,\textsuperscript{\rm 1}
    Junyang Wang\textsuperscript{\rm 1}
}
\affiliations{
    \textsuperscript{\rm 1} Beijing Jiaotong University\\
    yi.zhang@bjtu.edu.cn
}

\usepackage{bibentry}
\begin{document}

\maketitle

%%%%%%%%% ABSTRACT
\begin{abstract}
% fairness重要性
Deep learning models often learn to make predictions that rely on sensitive social attributes like gender and race, which poses significant fairness risks, especially in societal applications, e.g., hiring, banking, and criminal justice. 
% 或许可以不说这么复杂,直接说:在准确率之外,可信赖问题受到越来越多的关注.视觉识别中的公平性已经成为模型在真实世界大规模部署前必须解决的问题.
%previous methods
Existing work tackles this issue by minimizing information about social attributes in models for debiasing. However, the high correlation between target task and social attributes makes bias mitigation incompatible with target task accuracy.
%motivation 回顾了模型学习利用偏见属性进行决策是因为偏见属性有助于模型训练.构建替代
Recalling that model bias arises because the learning of features in regard to bias attributes (i.e., bias features) helps target task optimization, we explore the following research question: \emph{Can we leverage proxy features to replace the role of bias feature in target task optimization for debiasing?}
To this end, we propose \emph{Proxy Debiasing}, to first transfer the target task's learning of bias information from bias features to artificial proxy features, and then employ causal intervention to eliminate proxy features in inference.
% 说优势 核心思想是引导模型对偏见属性的学习放在人工代理特征上,相比于之前的去偏见方法,受益于我们去偏见不需要消除偏见属性因此我们可以兼容去偏见和目标任务准确率.
% The key idea of \emph{Proxy Debiasing} is to guide the model's learning of bias information on artificial proxy features, benefiting from our approach without preventing the learning of bias information, avoiding the fairness-accuracy paradox in previous methods.
The key idea of \emph{Proxy Debiasing} is to design controllable proxy features to on one hand replace bias features in contributing to target task during the training stage, and on the other hand easily to be removed by intervention during the inference stage. This guarantees the elimination of bias features without affecting the target information, thus addressing the fairness-accuracy paradox in previous debiasing solutions.
% 实验
We apply \emph{Proxy Debiasing} to several benchmark datasets, and achieve significant improvements over the state-of-the-art debiasing methods in both of accuracy and fairness.
\end{abstract}

%%%%%%%%% BODY TEXT
\section{Introduction}
\begin{figure}[t] 
    \centering
    \includegraphics[width=0.46\textwidth]{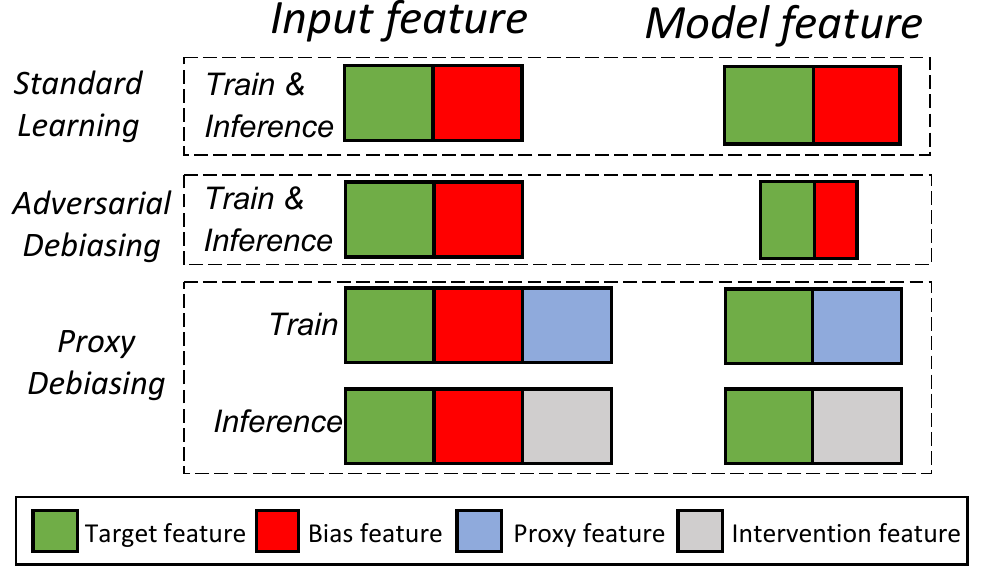}
    \caption{Illustration of the algorithmic bias problem, conventional debiasing method (Adversarial Debiasing) and the proposed proxy debiasing. Adversarial Debiasing method potentially removes partial target features in removing bias features, while our method uses proxy features to replace bias feature in learning of target task, which avoids unwittingly harming target features.
    }
    \label{fig:intro}
    \vspace{-3pt}
\end{figure}

Machine learning algorithms have achieved notable progress nowadays, and are increasingly being deployed in sensitive/high-stake environments to make important and life-changing decision, such as hiring, criminal justice, banking.
Nevertheless, there is growing evidence~\cite{buolamwini2018gender,wang2019racial,grother2019ongoing} that state-of-the-art models can perform potential discrimination based on bias attributes like gender and race, e.g., the popular COMPAS algorithm for recidivism prediction was found biased against black inmates and prone to make unfair sentencing decisions~\cite{lagioia2022algorithmic}.
These ethical issues have sparked a lot of research into fair machine learning.

% 最近的研究已经意识到不公平现象是因为标准训练的模型(图一中的第一行)继承了真实世界数据中的偏见,进而使模型在决策中不仅使用目标任务特征还学习了偏见特征(e.g.,gender features),正如图一中第一行所示.
Recent studies~\cite{locatello2019fairness,creager2019flexibly} realized that learning bias features (e.g., gender features) is one of the key factors causing unfairness. 
As shown in the first row in Figure~\ref{fig:intro}, models with standard learning inherit biased patterns in training data, and the learned decision rules thus depend on both target and bias features.
This realization spawned a lot of works to prevent models from learning bias features.
% 其中,朝着这个目标最直接的方法【】【】是直接删去训练数据中关于偏见特征的特征(e.g.,直接删除性别特征),然而,这只适用于结构化数据而不适用于视觉数据因为样本中目标特征和偏见特征是紧密耦合.
Towards this goal, the most direct method ~\cite{d2017conscientious} removes bias features from the training data (e.g., directly removing all information about gender). However, this only works for structured data and not for visual data, because the target features and bias features in the sample are tightly entangled.
In visual recognition, popular bias mitigation methods ~\cite{raff2018gradient,kim2019learning,jung2021fair} employ regularization terms to train models not to encode bias features of samples. 
Typical approaches, such as Adversarial Debiasing, adversarially train models to discriminate target attribute labels but fail to discriminate bias attribute labels. These works remove bias features in data representation, therefore enhancing output invariance to bias attribute.
% However, there is a limitation that they potentially remove useful target information for debiasing resulting in a loss of accuracy Because the target task information has a strong correlation with biased attribute information in biased training data. Also due to this correlation, the training process of the target task itself will promote the representation of bias attributes, resulting in limited debiasing effect. This game between target task learning and debiasing task leads to accuarcy-fairness paradox (cf., Figure 1(a)).
However, since target features and bias features have strong correlations in biased training data, these methods potentially remove partial target features and cause accuracy degradation in debiasing (cf., Figure~\ref{fig:intro}). Also due to this correlation, solving the target task inevitably promotes the representation of bias features, leading to limited debiasing effects. This zero-sum game between target task learning and debiasing task leads to the fairness-accuracy paradox.
% However, there is a limitation that they potentially remove 目标信息 and 导致了准确率的损伤 due 因为目标特征与偏见特征具有强的相关性in biased training data.同样由于这种相关性,目标任务本身训练的过程中又会促进对偏见属性的表示导致去偏见效果有限.这种目标任务学习和去偏见任务之间的博弈导致accuarcy-fairness paradox (cf., Figure 1(a)).

% 现有方法通过在训练中引入额外的公平性正则项来移除表征 模型对偏见特征的学习【】【】【】.典型的去偏见方法是对抗去偏见家族其中最典型的是和训练之外是否公平性策略介入训练阶段,第一类是在预测中使用直接修改模型输出阻止模型对偏见特征的学习
% To消除视觉识别中的偏见特征, many previous works (Wang et al. 2019b; Kim et al. 2019; Zhang, Lemoine, and Mitchell 2018)为目标任务的学习之外,以正则化项的形式 train models not to discriminate bias attribute labels.
% These works remove information related to protected attributes in data representation, thereby outputting invariant results in terms of the attributes. 

% 第三段,说目的 以及 idea
% In this paper,我们目的是:
In this paper, we aim to address the incompatibility between fairness and accuracy in debiasing.
% To this end,去偏见时我们需要在提出了一种避免与目标任务博弈的去偏见思路
To this end, we introduce \emph{Proxy Debiasing} that eliminates model bias without destroying the information of the target task.
As illustrated in Figure~\ref{fig:intro}, the key idea of \emph{Proxy Debiasing} is to use the artificial proxy features to replace the model's dependence on bias features in target task learning.
% , so that we does not need to impose additional fairness constraints on bias features to unwittingly harm target features.
% 前提是分布一致
The basic premise that the proxy feature can act as a proxy for the bias feature is that the proxy feature distribution should be consistent with the bias feature. To the end, we attach corresponding proxy features to samples with different bias attributes to satisfy distribution consistency, e.g., \emph{male} sample and \emph{female} sample are respectively attached with different proxy features in training, as shown in Figure~\ref{fig:datademo}.
Then, to eliminate the influence of proxy features on inference stage, we replace proxy features with intervention features based on causal intervention. Note that we do not need any prior information of sample, i.e., the intervention features imposed on all samples are the same.

We also conducted proxy effect analysis and reached a concluding observation: the model does not naturally learn bias information entirely from proxy features, and still learns bias information from bias features, which results in the inability to completely eliminate bias. 
% 我们
To solve this issue, we propose to maximize the contribution of proxy features to the target task to enhance the proxy effect of proxy features on bias features in the training of target task, which we call \emph{Active Proxy Debiasing}. For the contribution of proxy features, we borrow the idea of counterfactual explanations to measure. 
% 之后我们构造了额外观察发现,一对一不够,模型仍然会从偏见里学习,这是符合我们直觉的
% 因此我们又XXX提升模型对代理特征的依赖,以此模型在学习中会依赖代理
% 最后为了在测试阶段消除代理特征对模型决策的影响,我们基于因果干预将代理特征替换为了干预特征.note that 所有样本中的干预特征都是一致的,是偏见属性无关的和目标属性无关的.
Avoiding the zero-sum game between debiasing and target task learning in previous methods, our proxy debiasing method improves over previous methods on both fairness and accuracy. 

% 在使用代理特征代理消除偏见的过程中,整个方法工作像是一个普通的学习过程,我们没有使用额外的操作对模型表征进行公平性约束,因此我们叫做代理学习

We summarize our main contributions as follows:
\begin{itemize}
    \vspace{-3pt}
    % 我们提出顺应目标任务对数据中偏见信息的学习的去偏见idea,这避免了之前去偏见方法对目标任务信息带来的unintend损伤.
    \item We propose a novel debiasing method \emph{Proxy Debiasing} that employ proxy features to replace bias features in the target task's learning of bias information, which avoids the competition between target task and debias task in previous methods.
    % \vspace{-3pt}
    % 我们提出了proxy learning 方法.这个方法构建了能够提供给模型偏见信息的代理特征,并主动加强代理特征对目标任务贡献来提升代理特征对偏见特征的代理作用.结合测试阶段的因果干预,我们消除了代理特征的包含的偏见信息带来的影响.
    \item We introduce proxy effect enhancement that actively enhances the contribution of proxy features to the target task to improve the proxy effect of proxy features on bias features.
        % \vspace{-3pt}
    \item Extensive experiments demonstrate that our method significantly improves over baselines on both accuracy and fairness. The effectiveness on debiasing multiple bias attributes is also verified.
\end{itemize}

\begin{figure}[t] 
    \centering
    \includegraphics[width=0.46\textwidth]{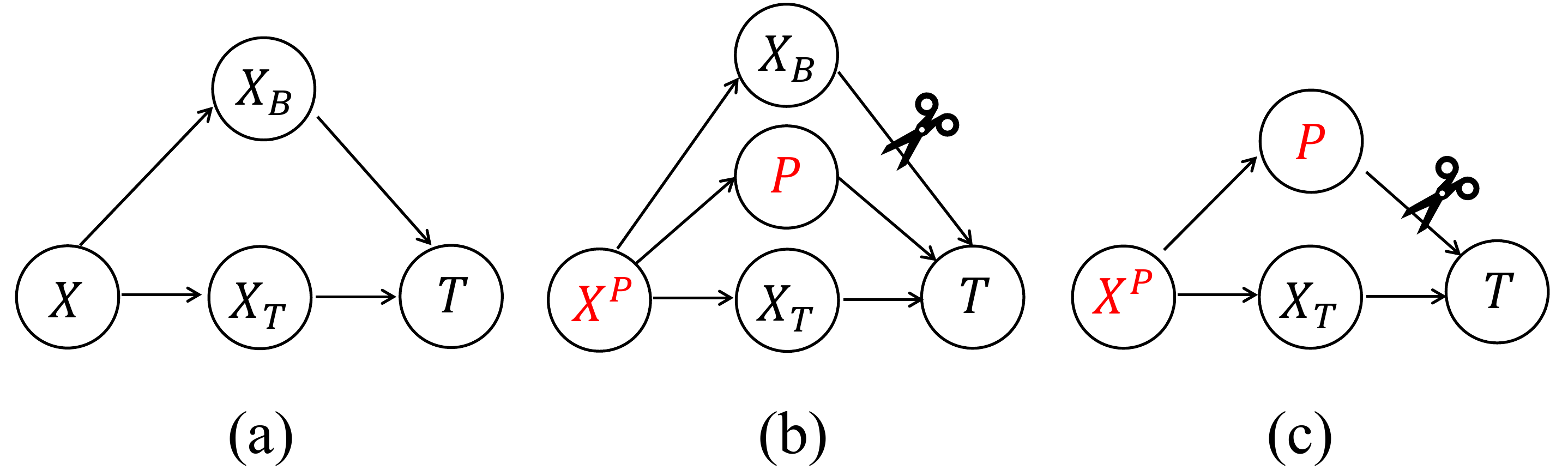}
    \caption{\textbf{The causal graph of the proposed model.} (a) The output $T$ of biased model is directly affected by the target feature $X_T$ and bias feature $X_B$ in input $X$. The training and inference stage of \emph{Proxy Debiasing} are illustrated in (b) and (c), respectively.
    }
    \label{fig:causalgraph}
    \vspace{-3pt}
\end{figure}

\section{Related Work}
% \subsection{Bias Mitigation}
\myparagraph{Bias mitigation.} 
% AI模型中的Gender 和 racial偏见已经被documented.
% Gender and race bias in AI models have been documented~\cite{yao2017beyond,zhao2017men}. 
% 以Equalodds(详细描述在sec4)为代表的偏见衡量方法揭示了深度学习exhibit social bias towards certain demographic groups.
% Bias measurement approach such as Equalodds reveal that AI models exhibit social bias towards certain demographic groups.
% 有很多工作已经被提出to tackle fairness problem in machine learning.
% Much research work has been done on mitigating model bias.
% 现存的公平性方法主要可以被分为三类根据消除方式.
Existing bias mitigation methods can be roughly divided into three families depending on the training pipelines they are applied to:
pre-processing methods~\cite{louizos2015variational,quadrianto2019discovering} refine dataset to mitigate the source of unfairness before training; in-processing methods~\cite{elkan2001foundations,jiang2020identifying} introduce fairness constraints into the training process; and post-processing methods~\cite{kamiran2012decision,pleiss2017fairness} adjust the prediction of models according to fairness criterion after training. Among them, in-processing methods have been the most studied due to no data recollection burden and significant accuracy drop.

% 大多数/典型 中处理 做法
Typical in-processing researches employ additional fairness constraint as regularization term for mitigating bias.
~\cite{zhang2018mitigating,wang2019balanced} enforce the model to produce fair outputs with adversarial training techniques by minimizing the ability of a discriminator to predict the bias attribute.
~\cite{kim2019learning} further minimizes the mutual information between representation and bias attributes to eliminate their correlations for debiasing. ~\cite{tartaglione2021end} devises a regularization term with a triplet loss formulation to minimize the entanglement of bias features. ~\cite{jung2021fair} tries to distill fair knowledge by enforcing the representation of student model to get close to that of the teacher model averaged
over the bias attributes.
% 存在的问题 然而,强相关性导致去偏见效果和
However, the high correlation between target task and bias attributes that exist in the data itself leads to the limited accuracy in debiasing.
Meanwhile, some methods try to convert the target task to not actively extract bias information. ~\cite{wang2020towards} trains different target models separately for each group in terms of bias attributes so that the target task does not attempt to rely on bias features. ~\cite{du2021fairness} trains classification head with neutralized representations, which discourages the classification head from capturing the undesirable correlation between target and bias information. 
% 然而这些方法仍然难以解决悖论,因为偏见存在数据本身
However, these methods still suffer from fairness-accuracy paradox due to the bias in the data itself.
% 在本文中,我们解决数据本身偏见通过构建偏见特征的代理特征.
In this paper, our method leverages artifacts features to proxy the bias feature in data itself for unraveling the fairness-accuracy paradox.

\myparagraph{Proxy features.} 
% 从偏见对特征学习的视角 背后机制是偏见特征代理了目标特征
~\cite{arpit2017closer} finds that models tend to learn features of easy patterns to proxy features of complex patterns in the data. 
Furthermore, the underlying mechanism of Shortcut~\cite{geirhos2020shortcut} can be seen as shortcut features proxying the intended features that humans want the model to use.
% 我们提出主动构建代理特征,解决原始特征难以定位的问题.
Inspired by this, we propose the concept of \emph{proxy feature} that actively replaces essence-related features with essence-independent features in model learning.
% 平行于偏见特征代理目标特征,我们提出目标特征代理偏见特征
% Parallel to the optimization of the target task leading to bias features to proxy target features, we propose to construct artificial features to proxy bias features.

\begin{figure*}[t]
    \centering
    \includegraphics[width=0.99\linewidth]{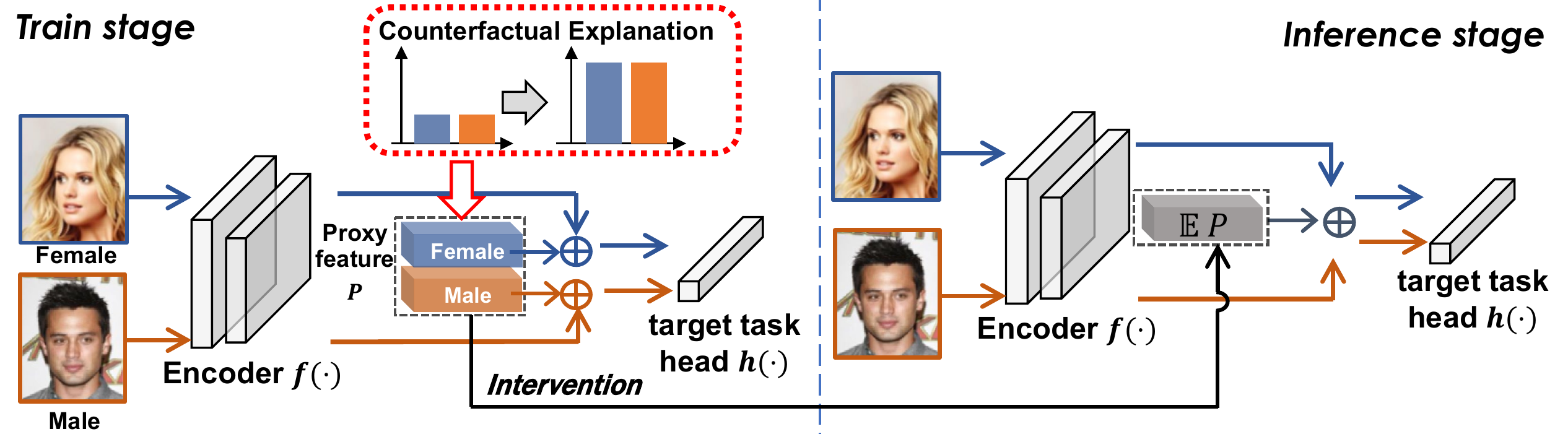}
    \caption{\textbf{Framework of Proxy Debiasing.} Active Proxy Debiasing contains additional proxy effect enhancement module (highlighted with \textcolor{red}{red} dash line) to Naive Proxy Debiasing.
    }
    \label{fig:overview}
    \vspace{-3pt}
\end{figure*}

\myparagraph{Causal intervention.} 
% 因果推断
Causal intervention has been widely used across many tasks to improve the robustness of deep learning models. Backdoor adjustment~\cite{pearl2014interpretation} is one of the most widely used implementations of causal intervention.
~\cite{yue2020interventional,zhang2020causal} leverage backdoor adjustment for eliminating the confounding factor in few-shot classification and weakly supervised segmentation. ~\cite{wang2020visual} employs backdoor adjustment to train feature extractors with commonsense knowledge. 
% 【】通过利用后门调整干预来训练具有常识知识的特征提取器,【】【】利用后门调整来训练鲁棒的小样本分类和弱监督分割.
% 与已有工作通过干预启发的训练不同,我们将干预应用于测试阶段来消除较好继承了训练数据的模型中的偏见.通过测试阶段的干预策略,我们可以避免在训练中消除偏见带来的目标任务和去偏见任务的零和博弈.
% Different from existing work to inspire training strategies with causal interventions, 
% we leverage causal intervention combined with the proxy feature strategy into the model inference stage to eliminate model bias.
Unlike existing work that applies causal intervention in the training phase, we apply causal intervention in the inference phase to remove the influence of proxy features.
%-------------------------------------------------------------------------

% \section{Fair intervention with proxy features}
\section{Methodology}
Visual recognition models, which are expected to only rely on target feature $X_T$ of the input $X$, are susceptible to making predictions based on the bias features $X_B$ of the data $X$, as illustrated in Figure~\ref{fig:causalgraph}(a). We aim to eliminate the model's dependence on $X_B$, i.e., the model output $T$ is independent of the bias features $X_B$, and prevent useful target features $X_T$ from being unintentionally corrupted in debiasing.

% \textsc{Definition 1} (\textsc{Fair visual recognition}). \emph{Given image data set $\mathcal{D}=\{\mathbf{x}_i,t_i,b_i\}_{i=1:N}$, where $\mathbf{x}_i\in \mathcal{X}$ denotes the $i^{th}$ image feature, $t_i\in \mathbb{T}$ denotes its target task label, $b_i\in \mathbb{B}$ denotes its 
% bias task label, fair visual recognition aims to learn an unbiased target task classifier satisfying that the prediction of target task label $\hat{t}$ is independent of the bias features $X_B$ and therefore independent of the bias labels: $P(\hat{t} = t_i|b;\theta) = P(\hat{t} = t_i;\theta)$. }

% Beyoud debiasing, we also aim to prevent useful target features $X_T$ from being unintentionally corrupted during debiasing.
% 受启发自proxy

Inspired by~\cite{arpit2017closer}, it is observed that among multiple features containing the same information, the model may only learn partial features, such as features with simple patterns.
We propose a \emph{Proxy Debiasing} method that performs debiasing in two stages: (1) Guide the model to preferentially use proxy features $P$ with simple patterns to learn bias information in target task training learning, so the model no longer needs to pay attention to bias features$X_B$ (cf., Figure~\ref{fig:causalgraph}(b)); (2) Introduce causal intervention mechanism in testing to eliminate the influence of proxy features on output $T$ (cf., Figure~\ref{fig:causalgraph}(c)).
% We propose Proxy Debiasing that uses proxy features to proxy the model's reliance on bias features in learning.
% , we propose Proxy Debiasing that aviod priovse
% The above data analysis section justifies the attribution of model bias from imbalanced data distribution and the potential of adversarial example in balancing data distribution. 

The direct way to realize \emph{Proxy Debiasing} is to attach pre-defined proxy features with simple patterns to original features, and the model then learns from this composite feature (see Figure~\ref{fig:overview}). This leads to the basic version of our solution, which we call \emph{Naive Proxy Debiasing} and will be introduced in the next subsection. 
% 然而,之后的分析表明模型并不会自然的优先学习代理特征.
However, subsequent analysis shows that proxy features with simple patterns do not naturally replace bias features.
% 去提升代理特征的优先级
To ensure the proxy effect of proxy features, we further propose to enhance the target task's attention to proxy features, which we call \emph{Active Proxy Debiasing} and as a complete version of our solution.

\begin{figure}[t] 
    \centering
    \includegraphics[width=0.43\textwidth]{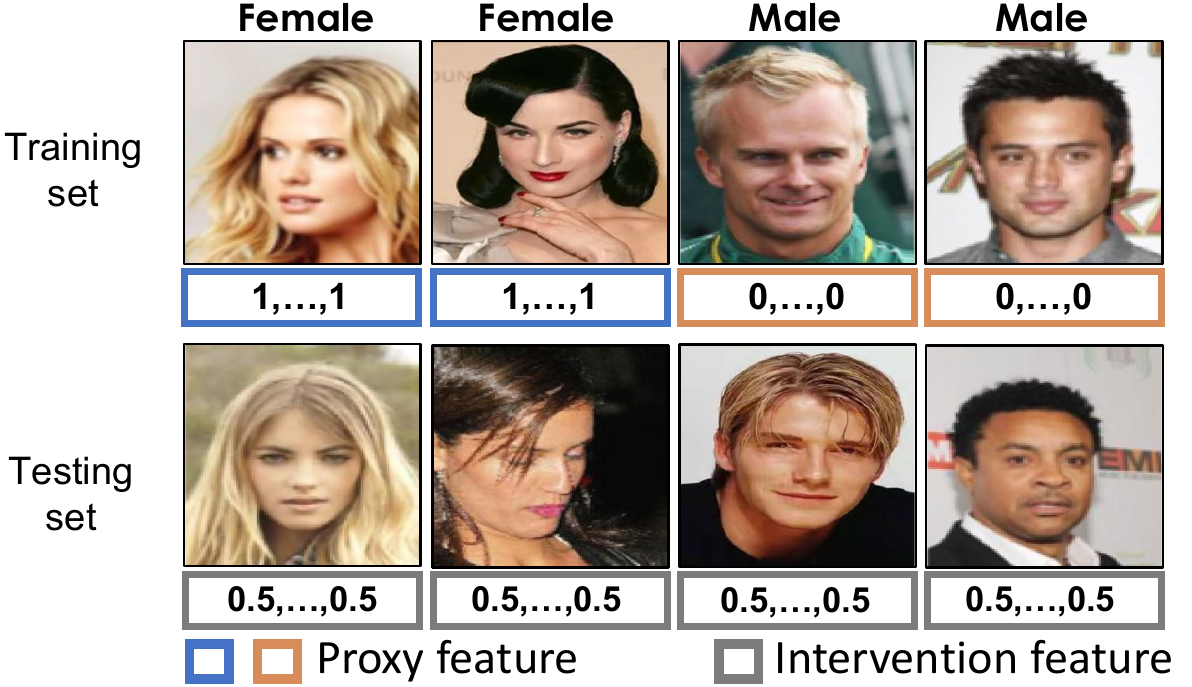}
    \caption{\textbf{Examples of Naive Proxy Debiasing for gender debiasing.} In training, images of different genders are assigned different proxy features, while all images are assigned the same intervention feature in testing.
    }
    \label{fig:datademo}
    \vspace{-3pt}
\end{figure}

\subsection{Naive Proxy Debiasing}
\myparagraph{Training with proxy features.} In fair visual recognition problem, input $x \in X$ is given two types of labels: target task attribute $t \in T$ and bias attribute $b \in B$.
% Under the motivation that use proxy features to replace 模型对偏见特征的依赖,我们需要构造与偏见特征分布一致的代理特征

% Under the motivation that to use proxy features to replace bias features, t
The premise of realizing proxying to bias features $X_B$ is that the proxy features $P$ should provide the same bias information as the bias features.
To this end, we construct proxy features that are consistent with the distribution of bias features.
As illustrated in Figure~\ref{fig:datademo}, due to the simple pattern is easy to learn, we preset all zeros or all ones vector as proxy features, and then utilize bias attribute label $b$ to select the corresponding proxy feature $p_b \in P$ and append it to the feature representation of the sample $x$.
% Note that we do not put the proxy feature as a patch in the image because it will occlude part of the image information.

Then, we train the target task in this composite data:
\begin{equation}\label{targetlearn}
\min  L_{\text {target}}(\{{x},p_b\}, t) 
\end{equation}
Where $\{{x},p_b\}$ is the model input that composite image $x$ and proxy feature $P_b$, $t$ is the target task label of $x$.

\myparagraph{Inference with intervention feature.}
% 在将替代后
Then, the dependence of trained models on bias information is based on proxy features $P$ rather than bias features $X_B$.
However, as shown in Figure~\ref{fig:causalgraph}, this introduces $P$ as a new source of bias in model inference:
\begin{equation}\label{equbias}
Pr(T \mid X)=Pr(T \mid X_T, P)
\end{equation}
% of different proxy features that correspond to different bias labels, such as male and female.

% A unbiased model should only be predict based on $X_T$ in image X, not on any bias information about $B$ including $P$.
And now, we need to remove the model bias brought about by proxy features $P$ in testing. Inspired by causal intervention use \emph{Do} operation to exclude the influence of confounder, we employ \emph{Do} operation to eliminate the effect of $P$ in the inference stage. Toward this, using the \emph{do} to prevent the introduction of causal effects of $P$ to $T$, the causal effects of target features $X_T$ to model output $T$ can be derived:
\begin{equation}
Pr(T \mid do(X_T))=\sum_{b}^{} Pr(T \mid X_T, p_b) Pr(p_b)
\end{equation}
Where $p_b$ is proxy feature corresponding to bias labels $b$. The underlying mechanism of \emph{Do} is to force $X_T$ to incorporate every $p_b$ fairly, subject to its prior $Pr(p_b)$. For low computation cost in testing, we replace $\sum_{b}Pr(Y\mid X_T, P_b) Pr(x_b)$ with $Pr(Y\mid X_T,\mathbb{E}_{\boldsymbol{b}} \left[p_b\right])$ due to \textbf{NWGM} linear approximation proved in ~\cite{xu2015show}:
\begin{equation}\label{inter}
Pr(Y \mid do(X_T))=Pr(Y \mid X_T, \mathbb{E}_{\boldsymbol{b}} \left[p_b\right]))
% \mathbb{E}_{\boldsymbol{z}}\left[g_{y}(\boldsymbol{z})
\end{equation}
Where $\mathbb{E}_{\boldsymbol{b}} \left[p_b\right]$ is mathematical expectation of $p_b$ subject to $b$, i.e., the mean of $p_{\boldsymbol{male}}$,
and is called intervention feature. 

Finally, we use intervention feature $\mathbb{E}_{\boldsymbol{b}} \left[p_b\right]$ to instead specific proxy features $p_b$ in inference. As shown in Figure~\ref{fig:datademo}, We use a vector of all 0.5. Note that another benefit of using intervention features is that we do not need to know the bias label of samples in inference. 

\begin{table}[t]

%   For each bias, we used four text templates to measure bias and calculate the aug.For each bias, we calculated using four text templates: Sentence w/o specify(Vanilla), Sentence w/ specify(Naive Proxy Debiasing), Phrase w/o specify(P$_\text{W/O}$) and Phrase w/ specify(P$_\text{W/}$)
  
%     \small
% 	\centering	
%     \aboverulesep = 0.46mm
%     \belowrulesep = 0.46mm
% \setlength\tabcolsep{5pt}
	\resizebox{0.47\textwidth}{!}
{
	\begin{tabular}	{l | l |  c  c  c  }
		\toprule
% 		\multirow{2}{*}{\multirow{2}{*}{VLPs}}
% 	  &  \multicolumn{3}{c}{TR}& \multicolumn{3}{c}{IR} \\
% 	 \midrule
	 \textbf{Task}&\textbf{Method}&\textbf{Acc.($\uparrow$)}&\textbf{ Bias($\downarrow$)}&\textbf{Counter@P}\\
    \midrule
	\multirow{2}{*}{{Blonde}}  &\textit{Vanilla}& 78.76 &40.82&- \\
	 &\textit{Naive PD}& \textbf{91.53}\footnotesize{(+12.77)} &\textbf{10.09}\textcolor{red}{\footnotesize{(-30.73)}}&\textcolor{red}{0.26} \\
	     \midrule
	\multirow{2}{*}{{Attractive}}  &\textit{Vanilla}& 76.71&26.09&- \\
	&\textit{Naive PD} & \textbf{77.25}\footnotesize{(+0.54)} &\textbf{24.06}\textcolor{red}{\footnotesize{(-2.03)}}&\textcolor{red}{0.01} \\ 
		\bottomrule
	\end{tabular}
 	}
 	
 \caption{The accuracy(in $\%$), model bias (described in Sec.4) and Counter@$P$ of Naive Proxy Debiasing (\textit{Naive PD}).}
 	
%  	 \vspace{-0.8ex}
% 	\caption
% 	{
% % 	\small	
% 		
% 	}
	\label{analysis}
    % \vspace{-0.7ex}
\end{table}

\subsection{Analysis of the Proxy Features}
% We examine the debiaisng performance of Naive Proxy Debiasing.
Taking face attribute recognition in CelebA~\cite{liu2015deep} as an example, we examine the gender debiasing performance of Naive Proxy Debiasing on Blonde and Attractive recognition tasks. As shown in Table~\ref{analysis}, compared to vanilla, we improve both fairness (lower model bias) and accuracy. However, neither model bias nor accuracy changes significantly in attractive recognition. 

Since we use the same configuration for both tasks, we conjecture that the reason for the above inconsistency in the two tasks might lie in the differences in the two tasks learning of proxy features. To test that, we look into the effect of proxy features on model output using counterfactual contrast, i.e., how the change of proxy features in test samples influences the model’s decision-making:
\begin{equation}\label{counterp}
Counter_{p}= \mathbb{E}_i \left| P(Y\mid x_i, p_{\boldsymbol{male}})-P(Y\mid x_i, p_{\boldsymbol{female}})\right|
\end{equation}
% Where $p_{m/f}$ is proxy feature regarding male or female.

% 解释结果 我们观察到去偏见效果好的blonde识别有着,继而模型不会将对偏见属性的依赖转移到代理特征上.
As reported in Table~\ref{analysis}, we note that in Naive Proxy Debaisng, for poor debiasing performance task (Attractive), the model has no significant dependence on proxy features(lower $Counter_{p}$). However, high debiasing performance task (Blonde) achieves significantly high $Counter_{p}$. 
This depicts that models do not always learn the simple pattern proxy features preferentially, even prioritizing learning of bias features, and therefore the proxy features $P$ cannot replace the model's reliance on bias features $X_B$.

\subsection{Active Proxy Debiasing}
The above observations suggest that simple proxy features may be trivial to target task,
we propose \emph{Active Proxy Debiasing} that adds additional \textbf{Proxy Effect Enhancement} module (red dash line in Figure~\ref{fig:overview}) to \emph{Naive Proxy Debiasing}.

% 代理特征的优先学习是。。。XXX
To this end, we guide the model actively to learn bias information from proxy features during target task training (Eq.~\ref{targetlearn}) so that the target task does not need to focus on bias features $X_b$.
% apply gradcam
% As illustrated in Figure 2, our model includes three trainable modules: proxy features, intervention features and classifier.

Specifically, instead of using preset proxy features, we use trainable proxy features to better satisfy the model's reliance on bias information in proxy features. 
% we use trainable proxy features for better meet the model on proxy features. 
Then, we borrow and revise the feature attribution strategy of counterfactual analysis~\cite{lang2021explaining,zhangcounter} to measure the importance of proxy features by counterfactually changing the proxy features:
\begin{equation}
\alpha^{c}=Y_c(x, p_{\tiny{b}})-Y_c(x, anchor)
\end{equation}
Where $\alpha^{c}$ indicates the importance of proxy features to target class $c$ (e.g., attractive or non-attractive), $Y_c(\cdot)$ denotes the logit output corresponding to class $c$, $p$ is the trainable proxy features corresponding to bias label $b$ of $x$. The $anchor$ is the preset counterfactual contrast point by randomly initializing, which is used as the counterfactual feature of $p$. 
% 我们应该XXX 然后其他

To reinforce the importance of proxy features for the model, we update proxy features $P$ and target task head $h$ to maximize the importance of proxy features with softmax normalization:
\begin{equation}\label{active}
\mathop{max}\limits_{P,h}  \frac{\exp \left(\alpha^{t}\right)}{\sum_{c=1}^{C} \exp \left(\alpha^{c}\right)}
\end{equation}
% Toward reinforce the importance of proxy features, we can update the proxy feature by maximize the contribution of proxy feature on target label:
% \begin{equation}
% \mathop{max}\limits_{p}\ { {Contri}_y}
% \end{equation}
Where $t$ is the target label of input $x$. By iteratively optimizing Eq.~\ref{targetlearn} and Eq.~\ref{active}, the trainable proxy features $P$ are guaranteed to continuously hold the proxy effect to bias feature $X_b$.
% 随着模型使用
% Same as Eq.~\ref{targetlearn}, we composite image $x$ and proxy feature $p$ as input to train target task.

\begin{algorithm}[tb]
\caption{Active Proxy Debiasing (\textit{Active PD})}
\label{alg:algorithm}
\begin{flushleft}

\textbf{Input}: {Training set $\mathcal{D} = \{(x_k, t_k, b_k)\}_{k=1}^{D} $}, the dimension of proxy features $\mathcal{M}$, the number of bias classes $\mathcal{N}$ \\
\textbf{Output}: Fair model\\
\end{flushleft}
% \textbf{Output}: \\
\begin{algorithmic}[1] %[1] enables line numbers
\STATE Initialization: Model parameter $\theta$, $\mathcal{N}$ $\mathcal{M}$-dimensional trainable proxy features $P$, one $\mathcal{M}$-dimensional counterfactual contrast point   $anchor$ 
\FOR{epoch 1,...,K}
% \For{\textup{minibatch} $\{(x_k, y_k, b_k)\}_{k=1}^N $}
\FOR{\textup{minibatch} $\{(x_k, t_k, b_k)\}_{k=1}^N $}
\STATE 
Step 1: Select $\{(p_{b_k})\}_{k=1}^N$ form $P$ according to $\{(b_k)\}_{k=1}^N$ 
\STATE Step 2: Update $\theta$ by minimizing target task loss (using Eq.~\ref{targetlearn})
\STATE Step 3: Get factual output $Y(x_k, p_k)$ of model
\STATE Step 3: Get counterfactual output $Y(x_k, anchor)$ of model with replacing ${(p_{b_k})}_{k=1}^N$ with $anchor$
\STATE Step 4: update trainable proxy features $P$ and target task head $h\in \theta$ to enhance proxy effect of proxy features for model (using Eq.~\ref{active})
\ENDFOR
\ENDFOR
\STATE Compute $\mathbb{E}_{\boldsymbol{b}} \left[p_b\right]$
\STATE Set the proxy feature of model to $\mathbb{E}_{\boldsymbol{b}} \left[p_b\right]$
\STATE \textbf{return} model $\theta$
\end{algorithmic}
\end{algorithm}

\myparagraph{Inference with intervention feature.}
Similar to Eq.~\ref{inter} in \emph{Naive Proxy Debiasing}, we also use intervention feature $\mathbb{E}_{\boldsymbol{b}} \left[p_b\right]$ to replace proxy feature. The difference is that proxy features $P$ here are trainable.

In summary, Algorithm 1 depicts the complete procedure of Active Proxy Debiasing (Active PD).

\section{Experiments}\label{Experiments}
% In this section, we first introduce fairness protocol, the datasets and baselines.
% We evaluate the effectiveness of our method, followed by extensive debiasing experiments in multiple bias attributes and ablation studies.

\subsection{Experiment Setup}
\myparagraph{Fairness metrics.} Many fairness criteria have been proposed including Statistical parity~\cite{feldman2015certifying}, Equal opportunity and Equalodds~\cite{hardt2016equality}.
Statistical parity requires that the probability of positive output of different groups is exactly equal, ignoring the label distribution of the test set itself.
Equal opportunity measure bias by comparing true positive rate between different groups. However, the fairness of positive and negative outputs is equally important, such as \emph{blonde} (Pos.) and \emph{non-blonde} (Neg.) in hair color recognition. Equalodds comprehensively considers fairness on all target labels as follows:
\begin{equation}
\frac{1}{|T|} \sum_{t} \left|\operatorname{Pr}_{b^{0}}(\tilde{T}=t \mid T=t)-\operatorname{Pr}_{b^{1}}(\tilde{T}=t \mid T=t)\right|
\end{equation}
where $T$ denotes target labels such as \emph{blonde}, $\tilde{T}$ denotes target outputs, and $b^{0}$ and $b^{1}$ represents different groups in terms of bias attributes such as \emph{male} and \emph{female}. 

\begin{table*}[th]
\centering
\resizebox{0.99\textwidth}{!}{
\begin{tabular}{c|cccccccccccccccccccc}
\toprule
\multirow{2}{*}{Method } &\multicolumn{2}{c}{T=\textit{a} , B=\textit{m}} && \multicolumn{2}{c}{T=\textit{bl} , B=\textit{m}} &&  \multicolumn{2}{c}{T=\textit{bn} , B=\textit{m}} &&  \multicolumn{2}{c}{T=\textit{a} , B=\textit{y}} &&\multicolumn{2}{c}{T=\textit{bl} , B=\textit{y}} && \multicolumn{2}{c}{T=\textit{bn} , B=\textit{y}}&& \multicolumn{2}{c}{\textbf{Avg.}} \\ \cmidrule[0.5pt]{2-3} \cmidrule[0.5pt]{5-6} \cmidrule[0.5pt]{8-9} \cmidrule[0.5pt]{11-12} \cmidrule[0.5pt]{14-15} \cmidrule[0.5pt]{17-18} \cmidrule[0.5pt]{20-21} 
& Acc.  & Bias &&  Acc.   &  Bias  && Acc.    &  Bias  &&  Acc.    &  Bias   && Acc.    &  Bias   &&  Acc.  &  Bias    &&  Acc.  &  Bias   \\ \cmidrule[0.5pt]{1-21} \morecmidrules\cmidrule[0.5pt]{1-21}
\textit{Vanilla} &  76.72 & 26.24 &&78.77 & 40.82 && 70.04 & 23.93  && 77.58 & 20.52&& 91.36  & 4.05 && 73.83 &  18.49  &&78.05&22.34   \\ \cmidrule[0.5pt]{1-21}
\textit{AdvDebias} & 77.54 & 11.56 &&79.24 & 33.44 &&  70.86 & 15.96  && 77.71 &  \textbf{10.48}&& 91.30 & 3.74 && 71.07 & 7.12  &&77.95& 13.71 \\ 
\textit{LNL} &  76.91 & 26.43 &&79.55 & 33.17 && 69.87 & 28.07  && 76.79 & 19.19&& 90.96 & 5.14 && 73.86 & 16.54 &&77.99&  21.42 \\ 
\textit{EnD} &  77.11 & 24.64 &&81.47 & 33.73 && 68.61 & 22.04  && 77.11 & 21.57&& 91.05 & 3.91 && 74.18 & 17.65&& 78.25& 20.59\\
\textit{MFD} &  77.22 & 20.17 &&79.85 & 38.84 && 71.07 & 28.86  && 77.31 & 22.00&& 90.92 & 5.16 && 75.39 & 16.12 &&78.62&21.85 \\ 
\textit{DI} &  77.53 & 23.01 &&91.44 & 7.76 && \textbf{73.03} & 15.89  && 77.69 & 17.17&& 91.05 & 4.66 && 73.98 & 10.64 &&80.78&13.18  \\
\textit{RNF} & 79.08 & 40.15 && 75.88 & 24.01 && 70.92 & 23.58  && 76.55 & 22.42&& 90.42 & 5.23 && 73.19 & 14.36 &&77.67&21.62  \\ %\cmidrule[0.5pt]{1-18}
\cmidrule[0.5pt]{1-21} 
% \textit{FSCL} & 11.5 & 79.1 && 13.0 & 79.1 && 7.0 & 82.1 && 6.4 & 83.8 && 3.8 & 82.7 && 1.8  & 82.0     \\ 
\textit{Active PD} & \textbf{79.70} & \textbf{7.33} && \textbf{92.02} & \textbf{4.97} && 72.11 & \textbf{2.56}  && \textbf{77.57} & 14.31&& \textbf{91.38} & \textbf{3.62} && \textbf{75.84} & \textbf{5.56} &&\textbf{81.43}& \textbf{6.36}  \\ \bottomrule
\end{tabular}
}
\caption{\textbf{The accuracy(in $\%$) and model bias(Equalodds) of models trained on CelebA.} Here T and B respectively represent target and bias attributes.  Here \textit{a}, \textit{bl}, \textit{bn}, \textit{m}, and \textit{y} respectively denote \textit{attractive}, \textit{blonde}, \textit{bignose}, \textit{male}, and \textit{young}.}
\label{table:celeba}
\end{table*}

\myparagraph{Datasets.} We evaluate the debiasing performance of Proxy Debiasing on \textbf{CelebA}~\cite{liu2015deep} and \textbf{UTKFace}~\cite{zhang2017age}. CelebA consists of more than 200,000 face images annotated with 40 binary attributes including two social concepts: \emph{Male} and \emph{Young}.  We set Male(\emph{m}) and Young(\emph{y}) as bias attributes, and select Attractive(\emph{a}), Blonde(\emph{bl}) and BigNose(\emph{bn}) as target attributes, due to vanilla trained models show unfairness in these target attributes. For UTKFace, we set Male(\emph{m}) and Ethnicity(\emph{e}) as target and bias attributes. For construct biased dataset, we truncate a portion of data to force the correlation $Pr(T|B)$ between target(T) and bias(B) attributes to be 0.9. For unbiased evaluation of the accuracy and fairness, the test set was constructed to have same number of samples for each target and each bias on both CelebA and UTKFace.

\myparagraph{Baselines.} We compare our Active Proxy Debiasing\textit{Active PD} against baselines such as: (1) Training DNN using cross-entropy loss without any debiasing technique(referred as \textit{Vanilla)}, (2) Adding regularization term regarding fairness constraints in the model optimization objective, including \textit{AdvDebias}~\cite{wang2019balanced} \textit{LNL}~\cite{kim2019learning}, \textit{End}~\cite{tartaglione2021end}, and \textit{MFD}~\cite{jung2021fair}, and (3) Controlling the contribution of bias features to target task toward fair generalization, including \textit{DI}~\cite{wang2020towards} and \textit{RNF}~\cite{du2021fairness}. 

\myparagraph{Implementation details.} 
We use ResNet-18~\cite{he2016deep} as the backbone network, and we initialize ResNet-18 with pretrained parameters. The vector dimension of proxy features is set to 100. For all baselines, we randomly sample the data with batchsize=128 and use Adam optimizer with learning rate=1e-3 and weight decay=1e-4.

\begin{table}[th]
\centering
\resizebox{0.47\textwidth}{!}{
\begin{tabular}{c|cccccccc}
\toprule
\multirow{2}{*}{Method } & \multicolumn{2}{c}{T=\textit{e} , B=\textit{m}} && \multicolumn{2}{c}{T=\textit{m} , B=\textit{e}}&& \multicolumn{2}{c}{\textbf{Avg.}} \\ \cmidrule[0.5pt]{2-3} \cmidrule[0.5pt]{5-6} \cmidrule[0.5pt]{8-9} 
& Acc.  & Bias &&  Acc.   &  Bias &&  Acc.   &  Bias  \\ \cmidrule[0.5pt]{1-9} \morecmidrules\cmidrule[0.5pt]{1-9}
\textit{Vanilla} & 84.98 & 24.14 && 88.53 & 15.70 &&86.74&18.42 \\ \cmidrule[0.5pt]{1-9}
\textit{AdvDebias} & 69.89 & 36.35 && 74.15 & 27.94 &&71.50&32.14   \\ 
\textit{LNL} & 85.01 & 22.28 && 87.31 & 19.13  &&86.16&20.70 \\ 
\textit{EnD} & 84.96 & 20.94 && 88.95 & 14.29  &&86.95&17.61 \\ 
\textit{MFD} & 84.19 & 24.79 && 87.75 & 17.74  &&85.97&21.26\\ %\cmidrule[0.5pt]{1-6}
\textit{DI} & 88.24 & 2.50 && 89.81 & 1.62&&89.02&2.06  \\ 
\textit{RNF} & 84.88 & 20.79 && 88.99 & 13.75  &&86.93&17.27\\ \cmidrule[0.5pt]{1-9}
% \textit{FSCL} & 11.5 & 79.1 && 13.0 & 79.1  \\ 
\textit{Active PD} & \textbf{90.34} & \textbf{0.99} && \textbf{91.10} & \textbf{0.96}  &&\textbf{90.72}&\textbf{0.97}   \\ \bottomrule
\end{tabular}
}

\caption{\textbf{The accuracy(in $\%$) and model bias(Equalodds) of models trained on UTKFace.} Here \textit{e} and \textit{m} respectively denote \textit{ethnicity} and \textit{male}.}

\label{table:utkface}
\end{table}

\subsection{Debiasing Performance Comparison}
We compare our Active Proxy Debiasing to state-of-the-art methods on both CelebA and UTKFace. Table~\ref{table:celeba} shows the accuracy and the model bias(EqualOdds) of models on diverse combinations of target and bias attributes of CelebA.
Averaging over all combinations, \emph{Vanilla} records the most severe model bias due to it is optimized to capture the statistical properties of training data without hindrance. Notably, our Active Proxy Debiasing(\textit{Active PD} for short) outperforms the previous methods on model bias and accuracy with a large margin.
% 公平性和准确率对其他方法的全面超越验证了我们使用代理去偏见防止目标任务学习和去偏见之间竞争的motivation
The fairness-accuracy compatibility validates our motivation that utilizes proxy features to prevent the zero-sum game between target task learning and debiasing, which results in a win-win for our method in terms of accuracy and fairness.
%  Moreover, the accuracy comparison between \emph{Vanilla} and Active Proxy Debiasing shows that there is no degradation in the accuracy of our method caused by an increase in fairness.
Furthermore, methods (\textit{AdvDebias}, \textit{LNL}, \textit{END} and \textit{MFD}) based on removing bias features from representation demonstrate insignificant debiasing performance. This suggests that the zero-sum game between target task learning and debiasing limits not only the accuracy but also the performance of debiasing.
% 其他方法取得一些效果
Other methods solve this issue by controlling the contribution of bias attributes to indirect debiasing. Notably, the accuracy and model bias of \textit{DI} is second only to ours. We conjecture that this is because \textit{DI} is an implicit form of avoiding competition between target tasks and debiasing, where separate target task classifiers are trained for each bias group.
\begin{table}[th]
\centering
\resizebox{0.47\textwidth}{!}{
\begin{tabular}{l|cc|cc|cc}
\toprule
\multirow{2}{*}{Method} & \multirow{2}{*}{Acc.} & \multirow{2}{*}{\textbf{Acc.@2}} & \multirow{2}{*}{Bias$_m$} & \multirow{2}{*}{\textbf{Bias$_m$@2}}& \multirow{2}{*}{Bias$_y$}& \multirow{2}{*}{\textbf{Bias$_y$@2}} \\
&&&&&&
\\
\cmidrule[0.5pt]{1-7} \morecmidrules\cmidrule[0.5pt]{1-7}
\textit{Vanilla} & 77.15 & 77.15 &26.24& 26.24 & 20.52& 20.52   \\ \midrule
\textit{AdvDebias} &\textcolor{red}{77.62}  & \textcolor{red}{64.21} &11.56& 48.28 & 10.48 & \textbf{7.98}    \\ 
\textit{LNL} & 76.85 & 76.85 &26.43& 28.83 & 19.19 & 23.91   \\ 
\textit{EnD} & 77.11 & 76.83 &24.64& 24.95 & 21.57 & 25.88  \\
\textit{MFD} & 77.26 & 77.20 &20.17& 26.23 & 22.00 & 21.42  \\ 
\textit{DI} & 77.61 & 77.41 &23.01& 24.48 & 17.17& 17.59  \\  
\midrule
\textit{Active PD} & 78.63 & \textbf{77.84} &7.33& \textbf{3.73} & 14.31  & 14.57  \\ \bottomrule
\end{tabular}
}
\caption{\textbf{The results of multiple biases debiasing.} Bias$_m$ and Bias$_y$ respectively represent the \emph{male} bias and \emph{young} bias in mitigation of corresponding single bias. Acc.$@$2, Bias$_m@$2 and Bias$_y@$2 denote accuracy, \emph{male} bias and \emph{young} bias, when both biases are eliminated simultaneously.} 
\label{table:multibias}
\end{table}

Table~\ref{table:utkface} summarizes the performance for different methods on UTKFace. The consistent observations with the above CelebA debiasing evaluation include: (1) \textit{Vanilla} records the most severe model bias. (2) Our \textit{Active PD} outperforms the previous methods in both model bias and accuracy (3) \textit{DI} performs second best in both model bias and accuracy due to \textit{DI} trains separate target task classifiers for each group. New observations include: For the case where the target attribute and the bias attribute are interchanged (T=\textit{e}/B=\textit{g} and T=\textit{g}/B=\textit{e}), our method can significantly eliminate the model bias, which shows that the debiasing ability of our method is independent of the setting of target attribute and bias attribute.

\subsection{Mitigation of Multiple Biases} 

% 之前去偏方法对有用目标任务信息的丢弃进一步限制了在多偏见场景下的去偏见,而我们的方法可以自然地应用与多偏见的去偏任务下,只需要构建多个artifacts来代理多种偏见.
Previous debiasing researches focus only on the debiasing of single bias.
To verify the debiasing performance on multiple biases, we modify these methods to debiasing for two biases. 
For \textit{Active Proxy Debiasing}, we construct two types of proxy features in each sample, representing male/female and young/old, respectively. For \textit{AdvDebias}, \textit{LNL} and \textit{MFD}, we incorporate fairness constraint in two bias attributes to eliminate two biases. For DI, we train 4 classifiers in terms of the male and young. And \textit{RNF} cannot be adapted to multiple bias attributes debiasing 
by simple modification. 

Taking \emph{male} and \emph{young} as two bias attributes, as shown in Table~\ref{table:multibias}, we report the debiasing performance in the attractive recognition task.
% 可以看出,我们的方法对于多个偏见的消除取得了显著偏见消除效果,其他方法的在去偏见时候难以均衡公平性和准确率.
The results show that Active Proxy Debiasing outperforms other methods in both accuracy and fairness, and the negligible model bias and accuracy gap between Multiple biases and single bias demonstrate our method can be applied to real-world multiple biases debiasing scenarios rather than only single bias debiasing simulations in laboratory settings.
Besides, other methods are not suitable for the mitigation of multiple biases. Particularly,
Mitigation of Multiple biases by \textit{AdvDebias} severely reduces accuracy compared to mitigation of single bias (from \textcolor{red}{77.62$\%$} to \textcolor{red}{64.21$\%$}).
Degradation of accuracy in \textit{AdvDebias} suggests that the fairness constraints on multiple biases have stronger competition with the learning of the target task, and result in the further fairness-accuracy paradox of mixing two biases. 
In contrast, our debiasing method avoids being an adversary for target task learning, thereby having the ability to eliminate multiple biases.

% \subsection{The Fairness-accuracy Compatibility} We compare the fairness-accuracy compatibility of XXX with other methods, and illustrate their fairness-accuracy curves for the two datasets in Figure 3. For regularization family methods, we vary their regularization weights to obtain the corresponding performance curves corresponding to different debiasing strengths.Our observations show that the fairness-accuracy compatibility of our method outperforms baselines with all debiasing strength choices.

\begin{figure}[t] 
    \centering
    \includegraphics[width=0.46\textwidth]{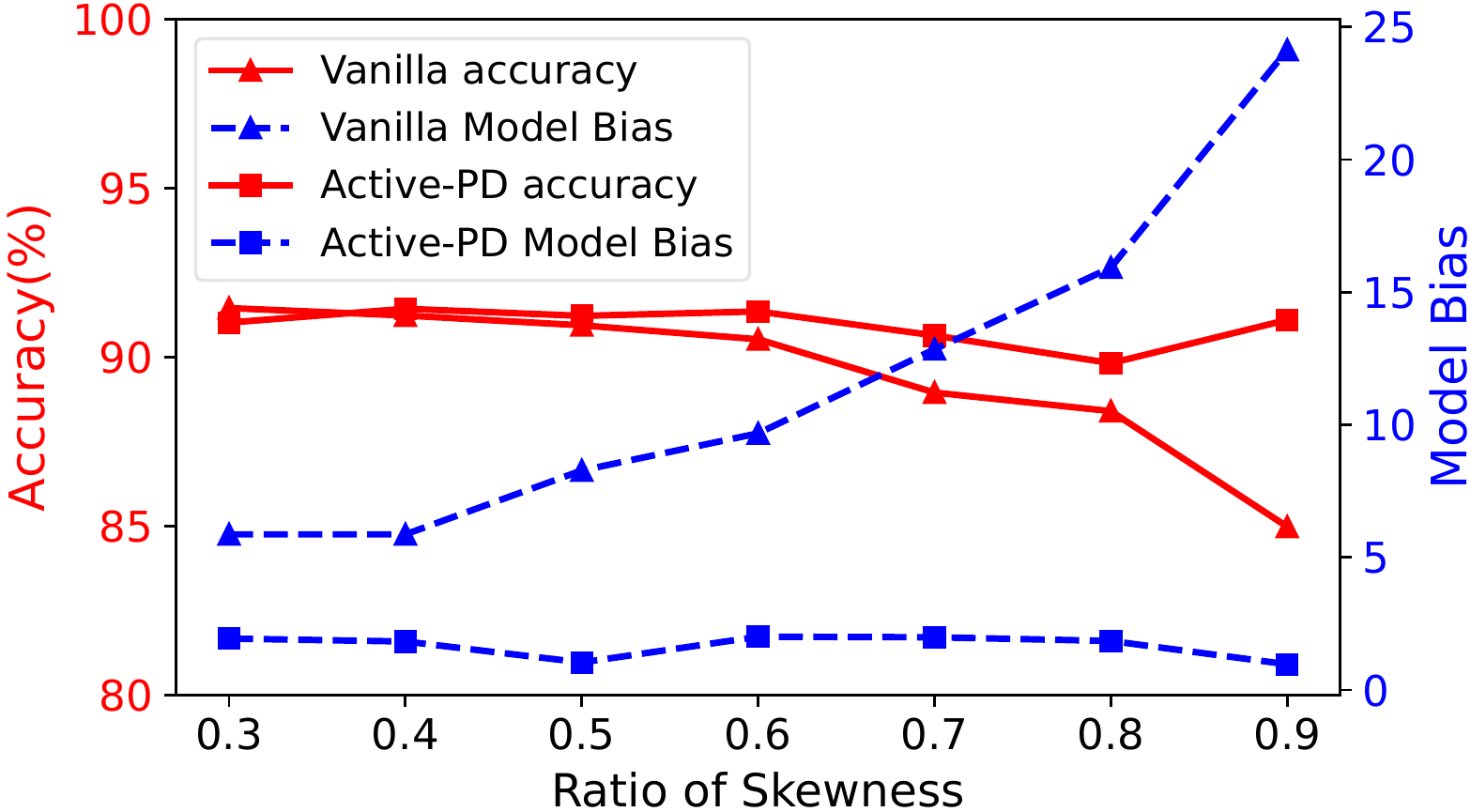}
    \caption{\textbf{Performance of Active Proxy Debiasing on UTKFace with varying ratio of $Pr$(\emph{male}$|$\emph{ethnicity}).}
    }
    \label{fig:databias}
    \vspace{-3pt}
\end{figure}

\subsection{Controlled Experiments in Various Data Bias} 
There are many cases of data imbalance in the real world. To simulate various data imbalances, we construct different datasets with various data imbalance ratio on UTKFace to evaluate the effectiveness and robustness of each method in various imbalance ratios. For more details, we truncate a portion of data to force the correlation $Pr(T|B)$ between target(male) and bias(ethnicity) attributes to be a list from 0.3 to 0.9 to simulate varying bias levels. 

In Figure~\ref{fig:databias}, we show the accuracy and model bias of vanilla and our method in various $Pr(T|B)$. It can be clearly noticed that both accuracy and fairness of Active Proxy Debiasing are basically not changed by the improvement of data bias $Pr(T|B)$, maintaining a greater advantage over Vanilla at all the intensities. And as $Pr(T|B)$ increases, accuracy gap between ours and vanilla becomes larger. This indicates that our method is robust to a variety of data bias scenarios, and our method can bring more gains in scenarios with more severe data bias.

\subsection{Qualitative Analysis with t-SNE Visualization}
\begin{figure}[t] 
    \centering
    \includegraphics[width=0.47\textwidth]{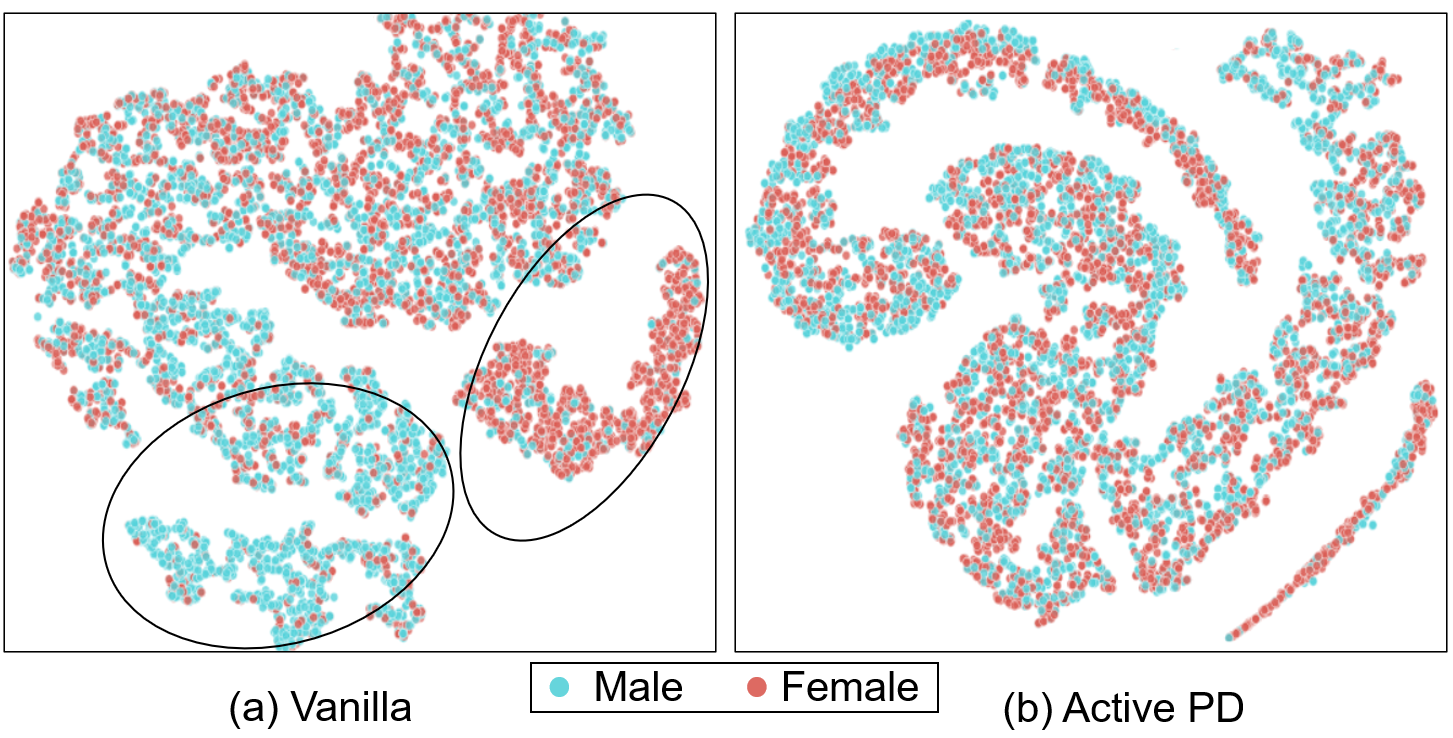}
    \caption{\textbf{Qualitative comparison using t-SNE visualizations.} 
    }
    \label{fig:tsne}
    \vspace{-3pt}
\end{figure}

To qualitatively investigate how \emph{Active Proxy Debiasing} successfully reduces the discrimination, we visualize
t-SNE embeddings of models trained with \emph{Vanilla} and \emph{Active Proxy Debiasing} in Figure~\ref{fig:tsne}(a) and (b).
The points of the figure are divided into two groups in terms of bias attribute (i.e., \emph{male} and \emph{female}), which are visualized in different colors.

In \emph{Vanilla}, the representation has separability for two bias attributes, especially in the oval region in Figure~\ref{fig:tsne}(a), suggesting that the models learn bias features attribute in target task learning. In contrast, in \emph{Active Proxy Debiasing}, the representation cannot be divided by bias attribute, that is, our method does not learn biased features in data. This visualization shows that our method mitigates the discrepancies between different groups.

% \subsection{Ablation Study}
\subsection{The Effectiveness of Proxy Effect Enhancement} 
We qualitatively validate the effectiveness of proxy effect enhancement module, i.e., optimizing proxy features and models to actively make the target task depend on proxy features so that the model can obtain sufficient bias information only from the proxy features. 

\begin{figure}[t] 
    \centering
    \includegraphics[width=0.47\textwidth]{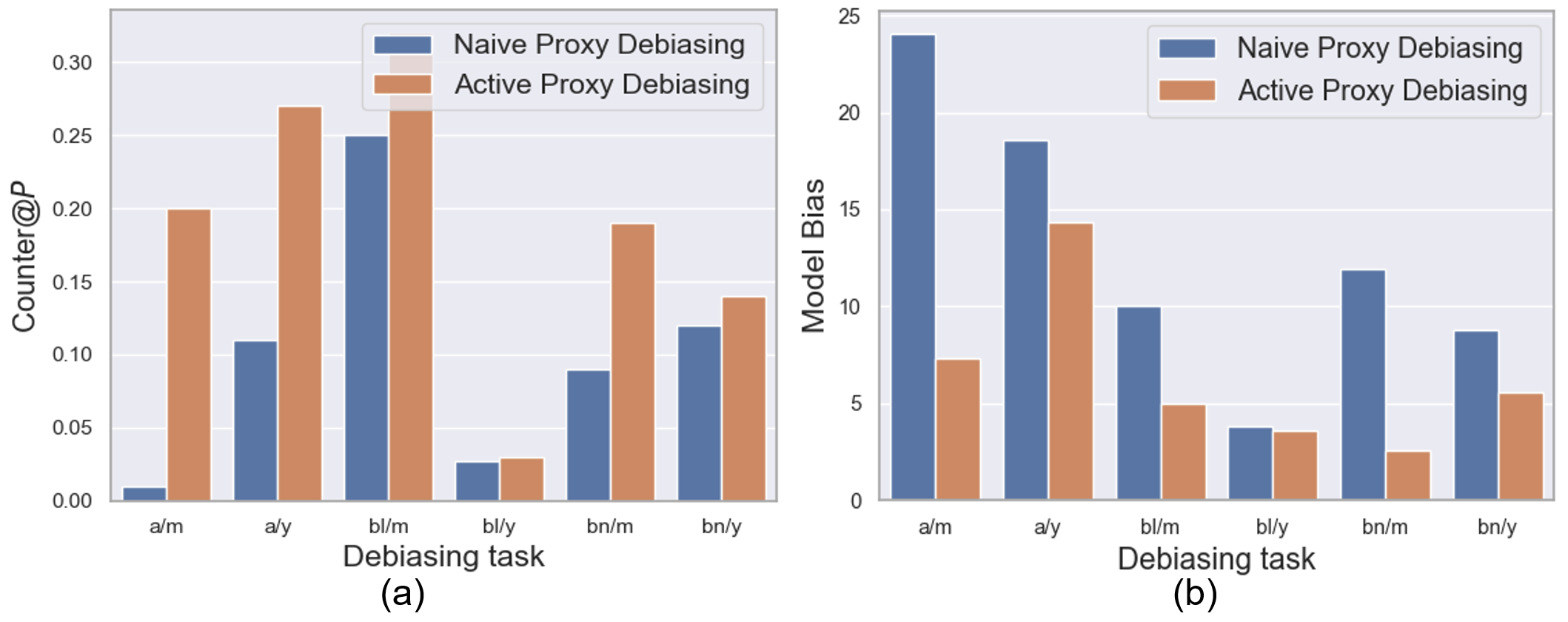}
    \caption{\textbf{The effect of proxy effect enhancement on CelebA.} It clearly shows that as Counter$@P$ is improved by the \textit{proxy effect enhancement} module in \textit{Active Proxy Debiasing} (left), the debiasing effect is further improved (right).}
    \label{fig:ablation}
    \vspace{-3pt}
\end{figure}

To validate, we use Counter$@P$ described in Eq.~\ref{counterp} to measure how dependent the model is on proxy features in the \emph{Naive} and \emph{Active Proxy Debiasing}.
In Figure~\ref{fig:ablation}(a), we report the Counter$@P$ of two methods in six debiasing tasks on CelebA. The plots demonstrate that \emph{Active Proxy Debiasing} significantly improves the dependence of the target task on proxy features (i.e., higher Counter$@P$ in all debiasing tasks). Also, we use model bias to measure how dependent the model is on real bias features in two Proxy Debiasing methods as shown in Figure~\ref{fig:ablation}(b). Combining the two plots, we can find that the stronger the model's dependence on proxy features, the more the model does not rely on real bias features. This indicates that when our active proxy strategy makes the target task already obtain enough bias information from the proxy features, the model will no longer use the real bias features as a decision-making basis.

\subsection{Proxy Feature Dimension Sensitivity }

\begin{figure}[t] 
    \centering
    \includegraphics[width=0.47\textwidth]{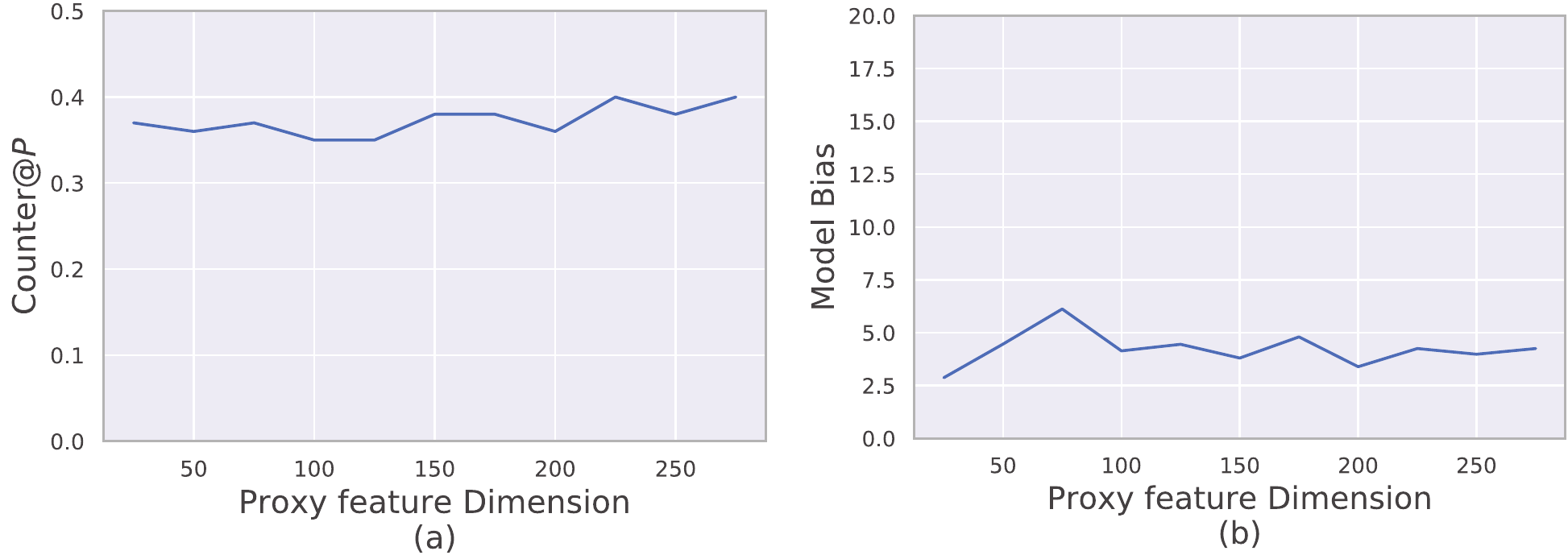}
    \caption{\textbf{Parameter sensitivity of Active Proxy Debiasing on Proxy Feature Dimension}. The proxy effect of proxy features (a) and debiasing performance (b) are basically unaffected by the choice of dimension. }
    \label{fig:sensitivity}
    \vspace{-3pt}
\end{figure}
We also conduct parameter sensitivity studies on proxy feature dimension on UTKFace.
We ran Active Proxy Debiasing at different proxy feature dimension settings, and Figure~\ref{fig:sensitivity} reports the Counter@$P$ (a) and debiasing performance (b). 
It clearly shows that the choice of proxy feature dimension has no significant effect on proxy effort (Counter@$P$) and debiasing.
This result shows that we can achieve debiasing using lower-dimensional proxy features, which means that debiasing can be achieved with lower computational cost in our method.
\section{Conclusion}
% 我们工作的目标是消除偏见的同时不破坏有用的目标任务信息.To this end,我们提出了对目标任务相关特征的因果干预来直接获得模型由目标信息作出的无偏预测,对于因果干预,我们使用了后门调整来实现.为了更有效的对偏见相关特征进行因果干预,我们提出了特征代理的方法使用人工特征代理模型中target task对偏见信息的应用.在多种数据集上的实验效果展示了我们方法significantly improves 之前的方法在准确率和去偏见上.
In this paper, we propose to employ proxy features to replace the target task's dependence on bias features towards visual debiasing. The proposed solution couples the operation of proxy effect enhancement and inference with the intervention feature, avoiding the use of bias features in training and proxy features in testing, respectively. The introduction of proxy features breaks through the fairness-accuracy paradox in previous methods, and the experimental results demonstrate its effectiveness in consistently improving fairness and accuracy.
% 此外对代理特征的使用还可以应用于其他问题？

\bibliography{Proxy_debias}

\begin{thebibliography}{35}
\providecommand{\natexlab}[1]{#1}

\bibitem[{Arpit et~al.(2017)Arpit, Jastrzebski, Ballas, Krueger, Bengio,
  Kanwal, Maharaj, Fischer, Courville, Bengio et~al.}]{arpit2017closer}
Arpit, D.; Jastrzebski, S.; Ballas, N.; Krueger, D.; Bengio, E.; Kanwal, M.~S.;
  Maharaj, T.; Fischer, A.; Courville, A.; Bengio, Y.; et~al. 2017.
\newblock A closer look at memorization in deep networks.
\newblock In \emph{International conference on machine learning}, 233--242.
  PMLR.

\bibitem[{Buolamwini and Gebru(2018)}]{buolamwini2018gender}
Buolamwini, J.; and Gebru, T. 2018.
\newblock Gender shades: Intersectional accuracy disparities in commercial
  gender classification.
\newblock In \emph{Conference on fairness, accountability and transparency},
  77--91.

\bibitem[{Creager et~al.(2019)Creager, Madras, Jacobsen, Weis, Swersky,
  Pitassi, and Zemel}]{creager2019flexibly}
Creager, E.; Madras, D.; Jacobsen, J.-H.; Weis, M.; Swersky, K.; Pitassi, T.;
  and Zemel, R. 2019.
\newblock Flexibly fair representation learning by disentanglement.
\newblock In \emph{International conference on machine learning}, 1436--1445.
  PMLR.

\bibitem[{d'Alessandro, O'Neil, and LaGatta(2017)}]{d2017conscientious}
d'Alessandro, B.; O'Neil, C.; and LaGatta, T. 2017.
\newblock Conscientious classification: A data scientist's guide to
  discrimination-aware classification.
\newblock \emph{Big data}, 5(2): 120--134.

\bibitem[{Du et~al.(2021)Du, Mukherjee, Wang, Tang, Awadallah, and
  Hu}]{du2021fairness}
Du, M.; Mukherjee, S.; Wang, G.; Tang, R.; Awadallah, A.; and Hu, X. 2021.
\newblock Fairness via representation neutralization.
\newblock \emph{Advances in Neural Information Processing Systems}, 34:
  12091--12103.

\bibitem[{Elkan(2001)}]{elkan2001foundations}
Elkan, C. 2001.
\newblock The foundations of cost-sensitive learning.
\newblock In \emph{International joint conference on artificial intelligence},
  973--978. Lawrence Erlbaum Associates Ltd.

\bibitem[{Feldman et~al.(2015)Feldman, Friedler, Moeller, Scheidegger, and
  Venkatasubramanian}]{feldman2015certifying}
Feldman, M.; Friedler, S.~A.; Moeller, J.; Scheidegger, C.; and
  Venkatasubramanian, S. 2015.
\newblock Certifying and removing disparate impact.
\newblock In \emph{proceedings of the 21th ACM SIGKDD international conference
  on knowledge discovery and data mining}, 259--268.

\bibitem[{Geirhos et~al.(2020)Geirhos, Jacobsen, Michaelis, Zemel, Brendel,
  Bethge, and Wichmann}]{geirhos2020shortcut}
Geirhos, R.; Jacobsen, J.-H.; Michaelis, C.; Zemel, R.; Brendel, W.; Bethge,
  M.; and Wichmann, F.~A. 2020.
\newblock Shortcut learning in deep neural networks.
\newblock \emph{Nature Machine Intelligence}, 2(11): 665--673.

\bibitem[{Grother, Ngan, and Hanaoka(2019)}]{grother2019ongoing}
Grother, P.; Ngan, M.; and Hanaoka, K. 2019.
\newblock Ongoing face recognition vendor test (FRVT) part 3: Demographic
  effects.
\newblock \emph{Tech. Rep. NISTIR 8280, National Institute of Standards and
  Technology}.

\bibitem[{Hardt, Price, and Srebro(2016)}]{hardt2016equality}
Hardt, M.; Price, E.; and Srebro, N. 2016.
\newblock Equality of opportunity in supervised learning.
\newblock \emph{Advances in neural information processing systems}, 29.

\bibitem[{He et~al.(2016)He, Zhang, Ren, and Sun}]{he2016deep}
He, K.; Zhang, X.; Ren, S.; and Sun, J. 2016.
\newblock Deep residual learning for image recognition.
\newblock In \emph{Proceedings of the IEEE conference on computer vision and
  pattern recognition}, 770--778.

\bibitem[{Jiang and Nachum(2020)}]{jiang2020identifying}
Jiang, H.; and Nachum, O. 2020.
\newblock Identifying and correcting label bias in machine learning.
\newblock In \emph{International Conference on Artificial Intelligence and
  Statistics}, 702--712. PMLR.

\bibitem[{Jung et~al.(2021)Jung, Lee, Park, and Moon}]{jung2021fair}
Jung, S.; Lee, D.; Park, T.; and Moon, T. 2021.
\newblock Fair feature distillation for visual recognition.
\newblock In \emph{Proceedings of the IEEE/CVF conference on computer vision
  and pattern recognition}, 12115--12124.

\bibitem[{Kamiran, Karim, and Zhang(2012)}]{kamiran2012decision}
Kamiran, F.; Karim, A.; and Zhang, X. 2012.
\newblock Decision theory for discrimination-aware classification.
\newblock In \emph{2012 IEEE 12th International Conference on Data Mining},
  924--929. IEEE.

\bibitem[{Kim et~al.(2019)Kim, Kim, Kim, Kim, and Kim}]{kim2019learning}
Kim, B.; Kim, H.; Kim, K.; Kim, S.; and Kim, J. 2019.
\newblock Learning not to learn: Training deep neural networks with biased
  data.
\newblock In \emph{Proceedings of the IEEE/CVF Conference on Computer Vision
  and Pattern Recognition}, 9012--9020.

\bibitem[{Lagioia, Rovatti, and Sartor(2022)}]{lagioia2022algorithmic}
Lagioia, F.; Rovatti, R.; and Sartor, G. 2022.
\newblock Algorithmic fairness through group parities? The case of
  COMPAS-SAPMOC.
\newblock \emph{AI \& SOCIETY}, 1--20.

\bibitem[{Lang et~al.(2021)Lang, Gandelsman, Yarom, Wald, Elidan, Hassidim,
  Freeman, Isola, Globerson, Irani et~al.}]{lang2021explaining}
Lang, O.; Gandelsman, Y.; Yarom, M.; Wald, Y.; Elidan, G.; Hassidim, A.;
  Freeman, W.~T.; Isola, P.; Globerson, A.; Irani, M.; et~al. 2021.
\newblock Explaining in style: Training a gan to explain a classifier in
  stylespace.
\newblock In \emph{Proceedings of the IEEE/CVF International Conference on
  Computer Vision}, 693--702.

\bibitem[{Liu et~al.(2015)Liu, Luo, Wang, and Tang}]{liu2015deep}
Liu, Z.; Luo, P.; Wang, X.; and Tang, X. 2015.
\newblock Deep learning face attributes in the wild.
\newblock In \emph{Proceedings of the IEEE international conference on computer
  vision}, 3730--3738.

\bibitem[{Locatello et~al.(2019)Locatello, Abbati, Rainforth, Bauer,
  Sch{\"o}lkopf, and Bachem}]{locatello2019fairness}
Locatello, F.; Abbati, G.; Rainforth, T.; Bauer, S.; Sch{\"o}lkopf, B.; and
  Bachem, O. 2019.
\newblock On the fairness of disentangled representations.
\newblock \emph{Advances in Neural Information Processing Systems}, 32.

\bibitem[{Louizos et~al.(2015)Louizos, Swersky, Li, Welling, and
  Zemel}]{louizos2015variational}
Louizos, C.; Swersky, K.; Li, Y.; Welling, M.; and Zemel, R. 2015.
\newblock The variational fair autoencoder.
\newblock \emph{arXiv preprint arXiv:1511.00830}.

\bibitem[{Pearl(2014)}]{pearl2014interpretation}
Pearl, J. 2014.
\newblock Interpretation and identification of causal mediation.
\newblock \emph{Psychological methods}, 19(4): 459.

\bibitem[{Pleiss et~al.(2017)Pleiss, Raghavan, Wu, Kleinberg, and
  Weinberger}]{pleiss2017fairness}
Pleiss, G.; Raghavan, M.; Wu, F.; Kleinberg, J.; and Weinberger, K.~Q. 2017.
\newblock On fairness and calibration.
\newblock \emph{Advances in neural information processing systems}, 30.

\bibitem[{Quadrianto, Sharmanska, and Thomas(2019)}]{quadrianto2019discovering}
Quadrianto, N.; Sharmanska, V.; and Thomas, O. 2019.
\newblock Discovering fair representations in the data domain.
\newblock In \emph{Proceedings of the IEEE/CVF conference on computer vision
  and pattern recognition}, 8227--8236.

\bibitem[{Raff and Sylvester(2018)}]{raff2018gradient}
Raff, E.; and Sylvester, J. 2018.
\newblock Gradient reversal against discrimination: A fair neural network
  learning approach.
\newblock In \emph{2018 IEEE 5th International Conference on Data Science and
  Advanced Analytics (DSAA)}, 189--198. IEEE.

\bibitem[{Tartaglione, Barbano, and Grangetto(2021)}]{tartaglione2021end}
Tartaglione, E.; Barbano, C.~A.; and Grangetto, M. 2021.
\newblock End: Entangling and disentangling deep representations for bias
  correction.
\newblock In \emph{Proceedings of the IEEE/CVF conference on computer vision
  and pattern recognition}, 13508--13517.

\bibitem[{Wang et~al.(2019{\natexlab{a}})Wang, Deng, Hu, Tao, and
  Huang}]{wang2019racial}
Wang, M.; Deng, W.; Hu, J.; Tao, X.; and Huang, Y. 2019{\natexlab{a}}.
\newblock Racial faces in the wild: Reducing racial bias by information
  maximization adaptation network.
\newblock In \emph{Proceedings of the IEEE/CVF International Conference on
  Computer Vision}, 692--702.

\bibitem[{Wang et~al.(2020{\natexlab{a}})Wang, Huang, Zhang, and
  Sun}]{wang2020visual}
Wang, T.; Huang, J.; Zhang, H.; and Sun, Q. 2020{\natexlab{a}}.
\newblock Visual commonsense r-cnn.
\newblock In \emph{Proceedings of the IEEE/CVF Conference on Computer Vision
  and Pattern Recognition}, 10760--10770.

\bibitem[{Wang et~al.(2019{\natexlab{b}})Wang, Zhao, Yatskar, Chang, and
  Ordonez}]{wang2019balanced}
Wang, T.; Zhao, J.; Yatskar, M.; Chang, K.-W.; and Ordonez, V.
  2019{\natexlab{b}}.
\newblock Balanced datasets are not enough: Estimating and mitigating gender
  bias in deep image representations.
\newblock In \emph{Proceedings of the IEEE/CVF International Conference on
  Computer Vision}, 5310--5319.

\bibitem[{Wang et~al.(2020{\natexlab{b}})Wang, Qinami, Karakozis, Genova, Nair,
  Hata, and Russakovsky}]{wang2020towards}
Wang, Z.; Qinami, K.; Karakozis, I.~C.; Genova, K.; Nair, P.; Hata, K.; and
  Russakovsky, O. 2020{\natexlab{b}}.
\newblock Towards fairness in visual recognition: Effective strategies for bias
  mitigation.
\newblock In \emph{Proceedings of the IEEE/CVF conference on computer vision
  and pattern recognition}, 8919--8928.

\bibitem[{Xu et~al.(2015)Xu, Ba, Kiros, Cho, Courville, Salakhudinov, Zemel,
  and Bengio}]{xu2015show}
Xu, K.; Ba, J.; Kiros, R.; Cho, K.; Courville, A.; Salakhudinov, R.; Zemel, R.;
  and Bengio, Y. 2015.
\newblock Show, attend and tell: Neural image caption generation with visual
  attention.
\newblock In \emph{International conference on machine learning}, 2048--2057.
  PMLR.

\bibitem[{Yue et~al.(2020)Yue, Zhang, Sun, and Hua}]{yue2020interventional}
Yue, Z.; Zhang, H.; Sun, Q.; and Hua, X.-S. 2020.
\newblock Interventional few-shot learning.
\newblock \emph{Advances in neural information processing systems}, 33:
  2734--2746.

\bibitem[{Zhang, Lemoine, and Mitchell(2018)}]{zhang2018mitigating}
Zhang, B.~H.; Lemoine, B.; and Mitchell, M. 2018.
\newblock Mitigating unwanted biases with adversarial learning.
\newblock In \emph{Proceedings of the 2018 AAAI/ACM Conference on AI, Ethics,
  and Society}, 335--340.

\bibitem[{Zhang et~al.(2020)Zhang, Zhang, Tang, Hua, and Sun}]{zhang2020causal}
Zhang, D.; Zhang, H.; Tang, J.; Hua, X.-S.; and Sun, Q. 2020.
\newblock Causal intervention for weakly-supervised semantic segmentation.
\newblock \emph{Advances in Neural Information Processing Systems}, 33:
  655--666.

\bibitem[{Zhang, Wang, and Sang(2022)}]{zhangcounter}
Zhang, Y.; Wang, J.; and Sang, J. 2022.
\newblock Counterfactually Measuring and Eliminating Social Bias in
  Vision-Language Pre-training Models.
\newblock \emph{CoRR}, abs/2207.01056.

\bibitem[{Zhang, Song, and Qi(2017)}]{zhang2017age}
Zhang, Z.; Song, Y.; and Qi, H. 2017.
\newblock Age progression/regression by conditional adversarial autoencoder.
\newblock In \emph{Proceedings of the IEEE conference on computer vision and
  pattern recognition}, 5810--5818.

\end{thebibliography}

\end{document}